\documentclass[11pt]{article}
\usepackage{anthology}
\usepackage{times}
\usepackage{array}
\usepackage{bm}
\usepackage{latexsym}
\usepackage{mathrsfs}
\usepackage{amsfonts} 
\usepackage{amssymb}
\usepackage{amsmath}
\usepackage{multirow}
\usepackage{bbding}
\usepackage{booktabs}
\usepackage{tabularx}
\usepackage{color}
\usepackage{graphicx}
\graphicspath{{images/} }
\DeclareGraphicsExtensions{.pdf,.png}

\usepackage[T1]{fontenc}
\usepackage[utf8]{inputenc}
\usepackage{microtype}
\usepackage{url}
  \let\oldurl\url
\usepackage{hyperref}
  
  \let\url\oldurl

\title{AIMA at SemEval-2024 Task 3: \\Simple Yet Powerful Emotion Cause Pair Analysis}

\author{\begin{tabular}{c}
Alireza Ghahramani Kure$^{\diamond}$, Mahshid Dehghani$^{\diamond}$, Mohammad Mahdi Abootorabi$^{\diamond}$, \\Nona Ghazizadeh$^{\diamond}$, Seyed Arshan Dalili$^{\diamond}$, Ehsaneddin Asgari$^{\mathsection}$
\end{tabular}\\$^{\diamond}$ NLP \& DH Lab, Computer Engineering Department, Sharif University of Technology \\ $^{\mathsection}$ Qatar Computing Research Institute, Doha, Qatar\\
\texttt{\{a.ghahramani, mahshid.dehghani, mahdi.abootorabi,}
\\
\texttt{nona.ghazizadeh, seyedarshan.dalili\}@sharif.edu}
\\
\texttt{easgari@hbku.edu.qa}
}

\begin{document}
\pagenumbering{gobble}
\maketitle
\begin{abstract}
The SemEval-2024 Task 3 presents two subtasks focusing on emotion-cause pair extraction within conversational contexts. Subtask 1 revolves around the extraction of textual emotion-cause pairs, where causes are defined and annotated as textual spans within the conversation. Conversely, Subtask 2 extends the analysis to encompass multimodal cues, including language, audio, and vision, acknowledging instances where causes may not be exclusively represented in the textual data. Despite this, our model addresses Subtask 2 using the same architecture as Subtask 1, focusing solely on textual and linguistic cues. Our architecture is organized into three main segments: (i) embedding extraction, (ii) cause-pair extraction $\&$ emotion classification, and (iii) post-pair-extraction cause analysis using QA. Our approach, utilizing advanced techniques and task-specific fine-tuning, unravels complex conversational dynamics and identifies causality in emotions. Our team, AIMA (MotoMoto at the leaderboard), demonstrated strong performance in the SemEval-2024 Task 3 competition ranked as the $10^{th}$ rank in subtask 1 and the $6^{th}$ in subtask 2 out of 23 teams. The code for our model implementation is available on \url{https://github.com/language-ml/SemEval2024-Task3}.
\end{abstract}

\section{Introduction}
\begin{figure*}[htb]
    \centering
    \includegraphics[width=\textwidth]{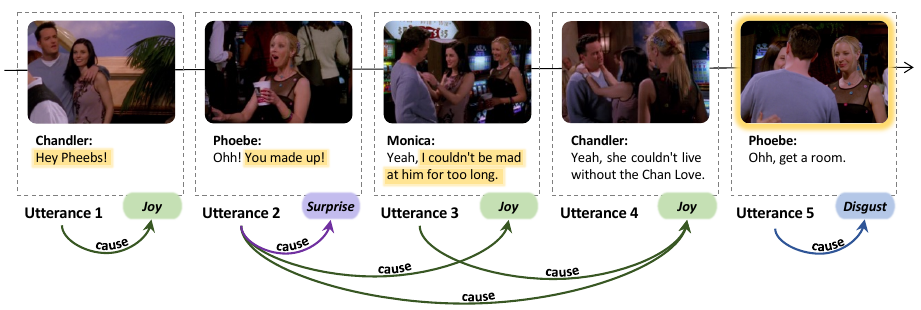}
    \caption{An example of the annotated conversation in ECF~\cite{wang2023multimodal} dataset, illustrating the multimodal nature of emotion causes. Each arc points from the cause utterance to the emotion it triggers. The cause spans have been highlighted in yellow.}
    \label{fig:dataset}
\end{figure*}

The task of Emotion-Cause Pair Extraction in Conversations holds significant importance in advancing the field of emotion analysis. Unlike previous endeavors that primarily focused on recognizing emotions, this task delves deeper into understanding the underlying causes behind emotional expressions within conversational contexts~\cite{wang2023multimodal}. Recognizing that emotions are conveyed not only through words but also through vocal intonations and facial expressions, the field has shifted towards multimodal emotion recognition. This move aims to understand how emotions are interwoven with text, sound, and visual cues in dialogue~\cite{wang2023multimodal}.
\\
The SemEval-2024 Task 3~\cite{wang-EtAl:2024:SemEval20244,wang2023multimodal,xia2019emotion} encompasses two subtasks aimed at extracting emotion-cause pairs in conversational contexts. Subtask 1 focuses on textual emotion-cause pair extraction, where causes are defined and annotated as textual spans within the conversation. In contrast, Subtask 2 broadens the analysis to incorporate multimodal cues, including language, audio, and vision. The task is based on the multimodal conversational emotion cause dataset ECF~\cite{wang2023multimodal}. Figure~\ref{fig:dataset} illustrates an example of the task and the annotated dataset.

\noindent In this paper, we introduce an approach based on a model architecture consisting of three key components: (i) embedding extraction, (ii) cause-pair extraction \(\&\) emotion classification, and (iii) cause extraction via QA post-pair detection. Utilizing advanced techniques and fine-tuning on specific datasets, our goal is to dissect complex conversational dynamics and pinpoint nuances that indicate emotional causality.
\\
Although our architecture supports multimodal data—including text, audio, and video through concatenations of the embeddings of these modalities using pretrained models—this study specifically harnesses textual data, as our primary focus is on addressing subtask 1. 

\section{Related Work}
This section provides an overview of two key areas in the field of emotion analysis: Emotion Recognition in Conversation and Emotion-Cause Pair Extraction in Conversations.
\\
\noindent\textbf{Emotion Recognition in Conversation:}
Emotion recognition in conversation, a burgeoning field, aims to decipher and understand the complex interplay of emotions within dialogues. ERC has seen significant advancements in recent years~\cite{kim2021emoberta, zheng2023facial}. These approaches have shown promising results on popular datasets such as IEMOCAP~\cite{busso2008iemocap} and MELD~\cite{poria2019meld}.
\\
EmoBERTa~\cite{kim2021emoberta} enhances RoBERTa~\cite{liu2019roberta} for emotion recognition in conversation (ERC) on datasets IEMOCAP~\cite{busso2008iemocap} and MELD~\cite{poria2019meld}, by incorporating speaker information and dialogue context. It preprocesses dialogues, representing them as sequences with speaker annotations and context segments. EmoBERTa extends RoBERTa to handle multiple segments and utilizes a linear layer with softmax nonlinearity for sequence classification.
\\
The FacialMMT~\cite{zheng2023facial} framework comprises two key stages. Initially, a pipeline method is employed to isolate the face sequence of the real speaker within each utterance. Following this, a multi-modal facial expression-aware emotion recognition model is applied. This model utilizes frame-level facial emotion distributions and incorporates multi-task learning to improve utterance-level emotion recognition. Experimental evaluations conducted on the MELD~\cite{poria2019meld} dataset validate the effectiveness of FacialMMT.
\\
\noindent\textbf{Emotion-Cause Pair Extraction in Conversations:}
The task of Emotion-Cause Pair Extraction in Conversations is pivotal for advancing our understanding of the nuanced interplay between emotions and their underlying triggers within dialogues, offering insights into human communication, cognition, and interpersonal dynamics.
\\
The paper~\cite{wang2023multimodal} introduces a baseline system, MC-ECPE-2steps, comprising two steps. Firstly, it employs multi-task learning to extract emotions and causes separately, utilizing word-level encoding and utterance-level encoders to derive representations specific to each. Secondly, it combines the predicted emotions and causes into pairs and employs BiLSTM and attention mechanisms to obtain pair representations. Subsequently, non-causal pairs are filtered out using a feed-forward neural network. Additionally, the system incorporates multimodal features from text, audio, and video modalities to enhance the extraction process. In addition to this approach, there exist other methodologies for Emotion-Cause Pair Extraction in Conversations~\cite{xia2019emotion, zheng-etal-2022-ueca}, some of which leverage question-answering techniques~\cite{nguyen2023emotioncause}.

\section{System Overview}
\begin{figure*}[htb]
    \centering
    \includegraphics[width=\textwidth]{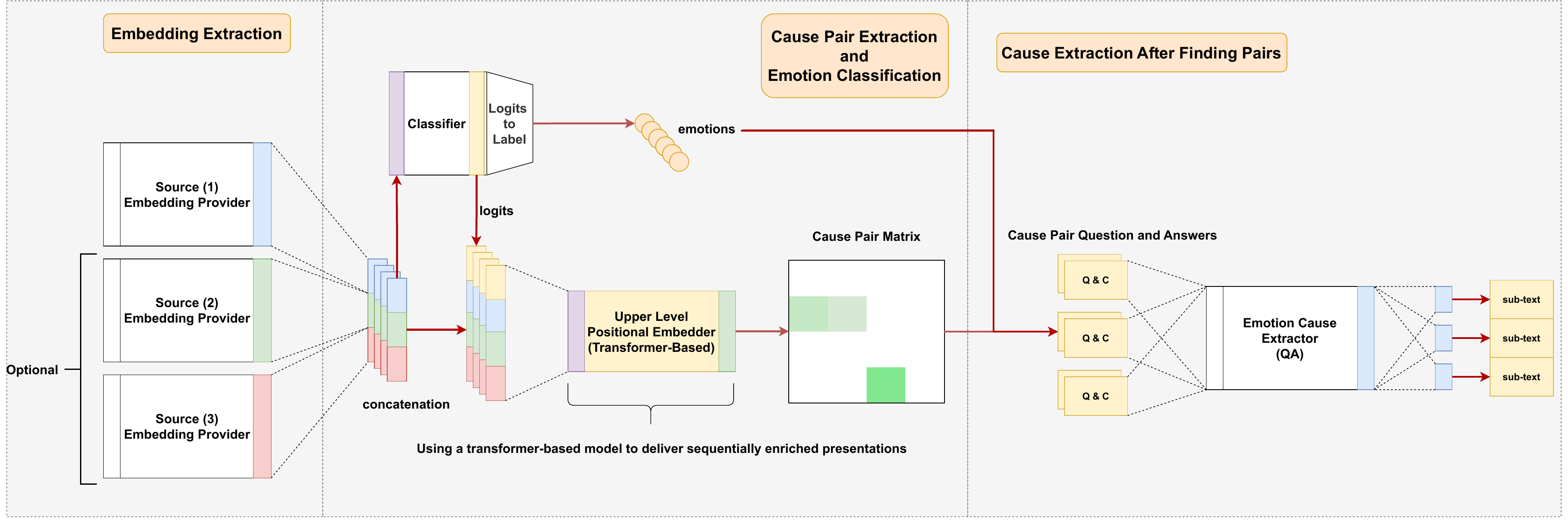}
    \caption{The schema of our proposed model for emotion-cause analysis, meticulously partitioned into three core segments: \textbf{Embedding Extraction}, \textbf{Cause Pair Extraction and Emotion Classification}, and \textbf{Cause Extraction After Finding Pairs}}
    \label{fig:model_schema}
\end{figure*}

Our model architecture, illustrated in Figure \ref{fig:model_schema}, is designed with the capacity to incorporate a diverse set of inputs from various sources such as text, video, and audio to perform emotion-cause analysis within conversational contexts.  However, for the purpose of addressing subtask 1, we specifically utilized textual data.

\noindent \textbf{Embedding Extraction and Emotion Classification:} 
In the Embedding Extraction phase, we leverage the EmoBERTa~\cite{kim2021emoberta} model specifically designed for text embedding. EmoBERTa's selection is based on its proven effectiveness in capturing the nuanced emotional dynamics inherent in conversational data, thereby facilitating precise emotion classification of the utterances. Additionally, it's noteworthy that EmoBERTa's emotion classification schema encompasses classes such as "neutral, joy, surprise, anger, sadness, disgust, and fear," mirroring the emotion categories present in the task dataset. This alignment ensures consistency in emotion classification across datasets. Moreover, we fine-tune EmoBERTa on the task dataset, further enhancing its ability to capture emotion-specific nuances within conversational utterances. Notably, the original model (before fine-tuning) achieves an accuracy of 67\% on the training data, indicating a good performance in emotion classification.

\noindent \textbf{Causality Matrix Extraction:} 
The embeddings of utterances, combined with logits from the classification task, are processed by a Transformer-based Encoder. This includes positional embeddings added to input vectors and a sequence of transformer encoder layers. The model's output, derived from the attention weights of the final layer, forms a causality matrix. This matrix highlights potential causal relationships within dialogue utterances, capturing the complex dynamics of conversation. The approach enriches data with emotion-specific insights, streamlining the identification of diverse emotion classes directly within the embeddings. In the following, the process of extracting the causality matrix is explained in detail.

\bigskip 

\noindent \textbf{Causality Matrix Extraction Process:}

\begin{enumerate}
    \item Initial combination of embeddings and logits:
    \begin{equation}
    combined = [s_1, s_2, s_3, logits]
    \end{equation}
    where \(s_1, s_2,\) and \(s_3\) are embeddings for an utterance, and \(logits\) are the output from the classification model \(M_c\), computed as \(logits = M_c([s_1, s_2, s_3])\).
    
    \item Application of dropout and addition of positional embeddings:
    \begin{equation}
    input = dropout(combined) + e_{pos}
    \end{equation}
    Here, \(e_{pos}\) represents the positional embeddings, which are added to the dropout-modified combined inputs to incorporate positional information into the sequence representation. Specifically, \(e_{pos}\) encodes the position of each utterance within the conversation, enriching the model's understanding of dialogue structure and the sequential context of each utterance.
    
    \item Generation of the causality matrix through the transformer encoder layers:
    \begin{equation}
    C_m = A_{N}(l^{encoder}_{1:N-1}(input))
    \end{equation}
    Here, \(l^{encoder}_i\) denotes the \(i\)-th transformer encoder layer, with \(N-1\) indicating that the input sequentially passes through all layers up to the \(N-1\)-th layer. \(A_{N}\) refers to the attention weights from of the \(N\)-th (last) encoder layer. The causality matrix, \(C_m\), is specifically derived from these attention weights applied to the output of the \(N-1\)-th layer, which has been processed by all preceding encoder layers and enhanced with positional embeddings. This matrix captures the causal interactions within the dialogue, as inferred from the attention mechanism of the transformer's final layer.

\end{enumerate}

\noindent \textbf{Question Generation for Causality Pairs:}
Following the emotion classification task, where emotions within the dialogue are identified, a causality matrix is created. For each emotion-cause pair detected in this matrix, the system generates a structured query to facilitate the extraction of the causal text segment. The prompt, constructed only for these detected pairs, follows the template:

"Which part of the text \{target\_utterance\} is the reason for \{speaker\}'s feeling of \{emotion\} when \{main\_utterance\} is said?"

\begin{figure}[ht]
    \centering
    \includegraphics[width=\linewidth]{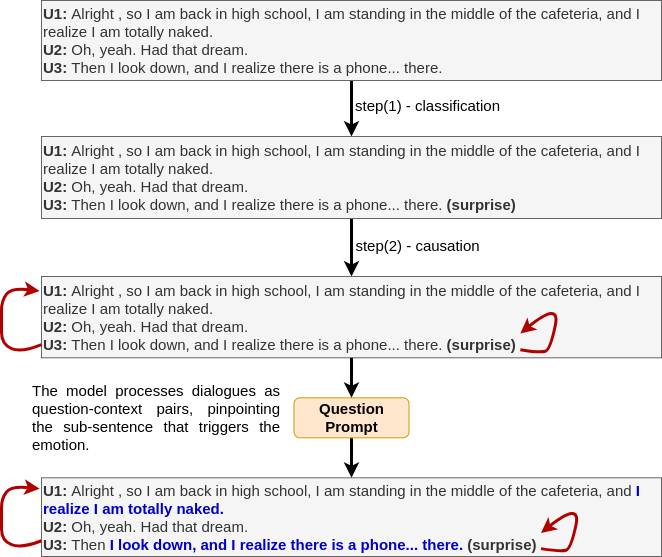}
    \caption{An example of the model's question-answering mechanism in action. After classifying emotions in the dialogue and creating the causality matrix, a question prompt is generated only for detected emotion-cause pairs. This diagram demonstrates the process of identifying the causative segment within the dialogue that led to the emotional response, with the causative text being highlighted in the context of the detected pairs.}
    \label{fig:qa_process}
\end{figure}

\noindent The Cause Extraction After Finding Pairs phase utilizes a question-answering model to interrogate the text, pinpointing exact sub-texts that substantiate the identified emotional triggers. (see Figure \ref{fig:qa_process}).

\noindent \noindent This study undertook a thorough evaluation of various question-answering (QA) models, uncovering areas where each model could be enhanced. Among the models examined, DistilBERT \cite{sanh2019distilbert} and BERT \cite{DBLP:journals/corr/abs-1810-04805} showed considerable promise for application within our research framework. Ultimately, we selected the deepset/deberta-v3-base-squad2, a pre-trained QA model, for our specific task requirements. This choice was informed by the model's foundation on the DeBERTa-v3-base architecture \cite{he2021debertav3} and its prior fine-tuning on the SQuAD2 dataset \cite{2016arXiv160605250R}, which includes both answerable and unanswerable questions. By further fine-tuning this model on our dataset, we ensured its proficiency in accurately extracting causal text segments from conversational contexts, a critical capability for our emotion-cause analysis.

\section{Experimental Setup}
\subsection{Dataset Preparation}
\noindent\textbf{Dataset Preparation for Attention Model:}
The dataset preparation for cause pair extraction and emotion classification procedure commenced with the loading of conversation data and emotion-cause pairs, accompanied by preprocessing steps tailored for model training. A custom dataset class facilitated the loading and processing of data, extracting essential details like conversation ID, utterances, and emotion-cause pairs. Subsequently, a collate function was employed to organize individual samples into batches suitable for model input, focusing solely on text and generating attention targets based on the presence of cause pairs within the textual data.

\noindent\textbf{Dataset Preparation for QA Model:}
The dataset preparation for subtext emotion cause extraction using question answering involved constructing samples for question answering by generating questions and contexts solely from text data. Each sample comprised a question formulated with a predefined prompt, the context concatenating all utterances from the conversation, and the answer containing the cause subtext. The dataset then underwent preprocessing to train the question-answering model, utilizing a pre-trained tokenizer to align tokenized inputs with the original text and determine the start and end positions of the answers within the textual context.

\subsection{Training}
\noindent\textbf{Training the Attention Model:}
The attention model was optimized using mean squared error loss and the AdamW optimizer with a learning rate of 1e-4.

\noindent\textbf{Training the QA Model:}
The QA model was trained over 25 epochs with a batch size of 8. 

\subsection{Evaluation Metrics}
Our models' performance was gauged using F1 scores across the six primary emotion categories, with additional emphasis on weighted averages to account for class imbalances. Subtask 1 evaluations incorporated both Strict Match and Proportional Match metrics to assess the accuracy of textual span identification for emotional causes.


\begin{table}[h]
\centering
\small 
\begin{tabular}{lccc}
\toprule
\textbf{Metric} & \textbf{Strict} & \textbf{Proportional} & \textbf{Weighted} \\
\midrule
Precision & 0.0217 & 0.2018 & 0.2779 \\
Recall & 0.0217 & 0.2081 & 0.2486 \\
F1-Score & 0.0217 & 0.2049 & 0.2584 \\
\bottomrule
\end{tabular}
\caption{Performance metrics for team AIMA (MotoMoto) in SemEval-2024 Task 3.}
\label{tab:results}
\end{table}

\section{Results}

\subsection{Quantitative Findings}
Our team, MotoMoto, participated in the SemEval-2024 Task 3 competition and secured the 10th rank in Subtask 1 and 5th rank in Subtask 2. The official metrics for our team's performance are as shown in Table \ref{tab:results} To explore the effectiveness of our approach, we compare it with the MC-ECPE-2steps~\cite{wang2023multimodal} method, which represents our baseline. The comparison is based on the weighted average F1 scores achieved by both approaches, as presented in Table \ref{tab:approach_comparison}. 

\begin{table}[htbp]
\centering
\small
\begin{tabular}{c|c}
\toprule
\textbf{Approach} & \textbf{Weighted-average F1} \\
\midrule
MC-ECPE-2steps & 0.3000 \\
-Audio & 0.2764 \\
-Video & 0.2993 \\
-Audio - Video & 0.2625 \\
\midrule
Ours & 0.2584 \\
\bottomrule
\end{tabular}
\caption{Comparison of Approaches with Baselines based on Weighted Average F1}
\label{tab:approach_comparison}
\end{table}

\subsection{Error Analysis}
Our investigation into the discrepancies between our system's predictions and the ground truth leveraged the detailed insights from the confusion matrix (Table \ref{tab:confusionMatrix}). The analysis underscores our emotion classification module's exceptional performance, notably in accurately identifying 'Neutral' and 'Joy' emotions with 4400 and 1576 correct instances, respectively. This substantiates our model's adeptness at recognizing emotions within conversations. Despite these strengths, the emotion-cause pair extraction component displayed variations, such as over or under-identification of causes compared to the ground truth annotations. Nevertheless, the precision of our model in identifying correct causes, as highlighted by specific successes in the confusion matrix, confirms its effectiveness in discerning emotions. These observations suggest that while our model excels in accurately identifying emotions, there is a valuable opportunity to refine the identification of causal factors within conversations for further improvement.

\begin{table}[ht]
\centering
\caption{Confusion Matrix for 13,619 dialogues. The model demonstrates no signs of overfitting, hence the entire train dataset is utilized to report this table.}
\label{tab:confusionMatrix}
\resizebox{\linewidth}{!}{ 
\begin{tabular}{|l|c|c|c|c|c|c|c|}
\hline
 & \rotatebox{90}{Neutral} & \rotatebox{90}{Joy} & \rotatebox{90}{Surprise} & \rotatebox{90}{Anger} & \rotatebox{90}{Sadness} & \rotatebox{90}{Disgust} & \rotatebox{90}{Fear} \\ \hline
Neutral & 4400 & 610 & 242 & 218 & 307 & 31 & 121 \\ \hline
Joy & 392 & 1576 & 136 & 82 & 70 & 19 & 26 \\ \hline
Surprise & 154 & 134 & 1380 & 77 & 34 & 17 & 44 \\ \hline
Anger & 168 & 180 & 192 & 823 & 88 & 71 & 93 \\ \hline
Sadness & 203 & 79 & 82 & 94 & 581 & 29 & 79 \\ \hline
Disgust & 83 & 34 & 41 & 77 & 25 & 143 & 11 \\ \hline
Fear & 70 & 36 & 42 & 24 & 35 & 8 & 158 \\ \hline
\end{tabular}}
\end{table}

\section{Conclusion}
Our investigation into emotion-cause pair extraction presents a paradigm shift towards simplicity and efficiency without compromising performance. By adopting a streamlined approach, we have demonstrated that high-impact emotion analysis does not necessarily require heavy computational resources or complex multimodal data integration. Our participation in the SemEval-2024 Task 3 competition has validated our methodology, securing commendable rankings and highlighting the efficacy of our model. The results underscore the potential of cost-effective solutions in the realm of emotion analysis, opening doors to wider applicability in resource-constrained environments. Looking forward, we aim to further optimize our model's efficiency and explore the integration of lightweight multimodal data processing techniques. This endeavor not only reinforces the viability of minimalist approaches but also sets a new benchmark for future research in emotion-cause analysis.

\bibliography{anthology}

\begin{thebibliography}{14}
\expandafter\ifx\csname natexlab\endcsname\relax\def\natexlab#1{#1}\fi

\bibitem[{Busso et~al.(2008)Busso, Bulut, Lee, Kazemzadeh, Mower, Kim, Chang, Lee, and Narayanan}]{busso2008iemocap}
Carlos Busso, Murtaza Bulut, Chi-Chun Lee, Abe Kazemzadeh, Emily Mower, Samuel Kim, Jeannette~N Chang, Sungbok Lee, and Shrikanth~S Narayanan. 2008.
\newblock Iemocap: Interactive emotional dyadic motion capture database.
\newblock \emph{Language resources and evaluation}, 42(4):335--359.

\bibitem[{Devlin et~al.(2018)Devlin, Chang, Lee, and Toutanova}]{DBLP:journals/corr/abs-1810-04805}
Jacob Devlin, Ming{-}Wei Chang, Kenton Lee, and Kristina Toutanova. 2018.
\newblock \href {http://arxiv.org/abs/1810.04805} {{BERT:} pre-training of deep bidirectional transformers for language understanding}.
\newblock \emph{CoRR}, abs/1810.04805.

\bibitem[{He et~al.(2021)He, Gao, and Chen}]{he2021debertav3}
Pengcheng He, Jianfeng Gao, and Weizhu Chen. 2021.
\newblock \href {http://arxiv.org/abs/2111.09543} {Debertav3: Improving deberta using electra-style pre-training with gradient-disentangled embedding sharing}.

\bibitem[{Kim and Vossen(2021)}]{kim2021emoberta}
Taewoon Kim and Piek Vossen. 2021.
\newblock \href {http://arxiv.org/abs/2108.12009} {Emoberta: Speaker-aware emotion recognition in conversation with roberta}.

\bibitem[{Liu et~al.(2019)Liu, Ott, Goyal, Du, Joshi, Chen, Levy, Lewis, Zettlemoyer, and Stoyanov}]{liu2019roberta}
Yinhan Liu, Myle Ott, Naman Goyal, Jingfei Du, Mandar Joshi, Danqi Chen, Omer Levy, Mike Lewis, Luke Zettlemoyer, and Veselin Stoyanov. 2019.
\newblock \href {http://arxiv.org/abs/1907.11692} {Roberta: A robustly optimized bert pretraining approach}.

\bibitem[{Nguyen and Nguyen(2023)}]{nguyen2023emotioncause}
Huu-Hiep Nguyen and Minh-Tien Nguyen. 2023.
\newblock \href {http://arxiv.org/abs/2301.01982} {Emotion-cause pair extraction as question answering}.

\bibitem[{Poria et~al.(2019)Poria, Hazarika, Majumder, Naik, Cambria, and Mihalcea}]{poria2019meld}
Soujanya Poria, Devamanyu Hazarika, Navonil Majumder, Gautam Naik, Erik Cambria, and Rada Mihalcea. 2019.
\newblock Meld: A multimodal multi-party dataset for emotion recognition in conversations.
\newblock In \emph{Proceedings of the 57th Annual Meeting of the Association for Computational Linguistics}, pages 527--536.

\bibitem[{{Rajpurkar} et~al.(2016){Rajpurkar}, {Zhang}, {Lopyrev}, and {Liang}}]{2016arXiv160605250R}
Pranav {Rajpurkar}, Jian {Zhang}, Konstantin {Lopyrev}, and Percy {Liang}. 2016.
\newblock \href {http://arxiv.org/abs/1606.05250} {{SQuAD: 100,000+ Questions for Machine Comprehension of Text}}.
\newblock \emph{arXiv e-prints}, page arXiv:1606.05250.

\bibitem[{Sanh et~al.(2019)Sanh, Debut, Chaumond, and Wolf}]{sanh2019distilbert}
Victor Sanh, Lysandre Debut, Julien Chaumond, and Thomas Wolf. 2019.
\newblock \href {https://api.semanticscholar.org/CorpusID:203626972} {Distilbert, a distilled version of bert: smaller, faster, cheaper and lighter}.
\newblock \emph{ArXiv}, abs/1910.01108.

\bibitem[{Wang et~al.(2023)Wang, Ding, Xia, Li, and Yu}]{wang2023multimodal}
Fanfan Wang, Zixiang Ding, Rui Xia, Zhaoyu Li, and Jianfei Yu. 2023.
\newblock Multimodal emotion-cause pair extraction in conversations.
\newblock \emph{IEEE Transactions on Affective Computing}, 14(3):1832--1844.

\bibitem[{Wang et~al.(2024)Wang, Ma, Xia, Yu, and Cambria}]{wang-EtAl:2024:SemEval20244}
Fanfan Wang, Heqing Ma, Rui Xia, Jianfei Yu, and Erik Cambria. 2024.
\newblock \href {https://aclanthology.org/2024.semeval2024-1.273} {Semeval-2024 task 3: Multimodal emotion cause analysis in conversations}.
\newblock In \emph{Proceedings of the 18th International Workshop on Semantic Evaluation (SemEval-2024)}, pages 2022--2033, Mexico City, Mexico. Association for Computational Linguistics.

\bibitem[{Xia and Ding(2019)}]{xia2019emotion}
Rui Xia and Zixiang Ding. 2019.
\newblock Emotion-cause pair extraction: A new task to emotion analysis in texts.
\newblock In \emph{Proceedings of the 57th Annual Meeting of the Association for Computational Linguistics}, pages 1003--1012.

\bibitem[{Zheng et~al.(2023)Zheng, Yu, Xia, and Wang}]{zheng2023facial}
Wenjie Zheng, Jianfei Yu, Rui Xia, and Shijin Wang. 2023.
\newblock A facial expression-aware multimodal multi-task learning framework for emotion recognition in multi-party conversations.
\newblock In \emph{Proceedings of the 61st Annual Meeting of the Association for Computational Linguistics (Volume 1: Long Papers)}, pages 15445--15459.

\bibitem[{Zheng et~al.(2022)Zheng, Liu, Zhang, Wang, and Wang}]{zheng-etal-2022-ueca}
Xiaopeng Zheng, Zhiyue Liu, Zizhen Zhang, Zhaoyang Wang, and Jiahai Wang. 2022.
\newblock \href {https://aclanthology.org/2022.coling-1.613} {{UECA}-prompt: Universal prompt for emotion cause analysis}.
\newblock In \emph{Proceedings of the 29th International Conference on Computational Linguistics}, pages 7031--7041, Gyeongju, Republic of Korea. International Committee on Computational Linguistics.

\end{thebibliography}
\bibliographystyle{acl_natbib}

\end{document}